\documentclass[twocolumn]{article}
\usepackage{graphicx}
\graphicspath{{./figs/}}
\usepackage{todo}
\usepackage{xcolor}
\usepackage{amsmath}
\usepackage{multirow}
\usepackage{booktabs}
\usepackage{subfig}
\usepackage{bm}
\usepackage{hyperref}
\usepackage{tikz}
\usepackage{threeparttable}

\newif\ifanonim
\anonimfalse
\ifanonim
\def\repoURL{\url{(URL to be provided in the non-anonymous version)}}
\def\repoURLdata{\url{(URL to be provided in the non-anonymous version)}}
\def\acknowledge{Acknowledgements omitted in the anonymized version.}
\author{Anonimized}
\institute{Anonimized\\ \email{anonim@anonim.edu}}
\else
\def\repoURL{\url{https://github.com/lquirosd/ODDLA}}
\def\repoURLdata{\url{https://github.com/lquirosd/ODDLA/tree/master/data}}
\def\acknowledge{%
Work partially supported by 
the Universitat Polit\`ecnica de                                                                                
Val\`encia under grant FPI-II/900, 
by EU JPICH project ``HOME {\textendash} History Of Medieval Europe'' (PEICTI Ref. PCI2018-093122),                                                                            
by the Generalitat Valenciana under the EU-FEDER Comunitat                                                                                                 
Valenciana 2014-2020 grant  IDIFEDER/2018/025 ``Sistemas de fabricaci\'on                                                                                  
inteligente para la ind\'ustria 4.0'',                                                                                                                     
and by Ministerio de Ciencia, Innovaci\'on 
y Universidades project DocTIUM (Ref.                                                                           
RTI2018-095645-B-C22).
}

\author{Lorenzo Quir\'os, Enrique Vidal\\
        \small PRHLT Research Center, Universitat Polit\`ecnica de Val\`encia, Camino de Vera, s/n, 46022 Valencia, Spain\\                  
        \small \textless loquidia,evidal\textgreater @prhlt.upv.es
        }%

\date{}
\fi

\title{Evaluation of a Region Proposal Architecture for Multi-task Document Layout Analysis}
\providecommand{\keywords}[1]
{
      \small    
        \textbf{\textit{Keywords---}} #1
    }

\begin{document}
\maketitle              %
\begin{abstract}
Automatically recognizing the layout of handwritten documents
is an important step towards useful extraction of information
from those documents.  The most common application is to feed
downstream applications such as automatic text recognition and
keyword spotting; however, the recognition of the layout also
helps to establish relationships between elements in the document
which allows to enrich the information that can be extracted.
Most of the modern document layout analysis systems are designed
to address only one part of the document layout problem, namely:
baseline detection or region segmentation.  In contrast, we
evaluate the effectiveness of the
Mask-RCNN architecture to address the problem of baseline detection and
region segmentation in an integrated manner. We present experimental results
on two handwritten text datasets and one handwritten music dataset. The analyzed
architecture yields promising results, outperforming state-of-the-art techniques
in all three datasets.

\end{abstract}

\section{Introduction}
\label{sec:intro}
Document Layout Analysis (DLA) is a well-established area of research that aims 
to extract the structure with which the information that the writer wanted to 
convey was recorded on paper. 
This includes the lines of text, the structural 
groups into which the lines are organized and their role in the document (paragraph,
marginalia, title, etc.), the regions of non-textual information (illustrations, 
musical staff, graphs, etc.), and finally, the relationships between all the elements
in the document (reading order, linked references, etc.).

It is important to notice that even with a perfect automatic text recognition 
system (TSR, which stands for OCR, HTR, or HMR)~\cite{Graves09,romero2011,Bluche15,sanchez2019,martinek2020} 
the information that can be extracted automatically from 
the document without a proper DLA system is very limited. For instance, a perfect
text line level transcription only is completed when all text lines are grouped 
into their respective structural groups (e.g. paragraphs) in reading order.

What is more, most state-of-the-art TRS are designed 
to handle only one line of text at a 
time~\cite{Bluche15,puigcerver2018,toselli2016}, so a DLA system must pre-process
the images in order to split the document into a set of text lines. Even in the 
case of the few modern TRS that are able to process a full paragraph at a
time~\cite{huang2019,carbonell2019,wigington2018,bluche2017} a DLA system is 
needed
to pre-process the document and split it into those 
paragraphs. 
If the document layout is simple enough, some of these systems
might be able to handle full pages without any previous DLA, but
they would fail to handle documents with complex layouts.
Consequently, most modern DLA systems aim to
address the text line segmentation
problem (or baseline detection)~\cite{gruning2018,ares2018,lquiros2018} in order 
to feed a TRS an obtain the
corresponding transcript. Although it is an important step in the process, 
normally
it is not enough to arrange the text lines in their structural groups and move 
towards a complete transcription process. 

In this sense other works propose to train new DLA systems to segment the 
document into a set of structural groups~\cite{ares2018,gruning2018}, so 
text-lines can be grouped later on. 
On the other hand, in~\cite{lquiros2018,lquiros2019} an integrated model is 
proposed to handle both tasks at the same time so the text lines become already
grouped as part of the process. Despite the good results reported, the main 
drawback of both kind of systems is the inability to properly handle 
overlapping regions.

In general DLA sub-tasks, such as text line and region segmentation, can be seen as 
an object segmentation problem~\cite{haralick1985}, where each region of interest
in the document (textual and non-textual) is a different object in the 
image. For example, illustrations, paragraphs, text-lines and marginalia. Hence, 
some works draw upon the deep learning systems developed for general object segmentation
to improve text line and region segmentation. For instance, in~\cite{zhong2019,saha2019}
a region proposal neural network is used for region segmentation on printed
documents, while in~\cite{prusty2019,studer2019} similar systems are used for
region segmentation and text line segmentation on handwritten text documents.
However, to the best of out knowledge, there is no analysis of this kind of 
architectures on the multi-task setting of DLA in complex documents.

In this work, we analyze the use of a region proposal network architecture 
(Mask-RCNN~\cite{he2017})
to address the problem of multi-task DLA (integrated baseline detection and region 
segmentation) on complex documents. Several training 
scenarios are analyzed, both in handwritten text and music documents. 

The paper is organized as follows. In Sec.~\ref{sec:prop} we describe the system
under analysis in this work. Section~\ref{sec:exp} describes the experimental 
setup adopted, datasets and model hyper-parameters. In Sec.~\ref{sec:results} we
present and discuss out results, while final remarks and future research 
directions are discussed in Sec.~\ref{sec:conc}. Finally, in Sec.~\ref{sec:rep} 
details to reproduce our experiments are provided.

\section{Multi-Task Document Layout Analysis}
\label{sec:prop}

Even though DLA may be considered a much broader and complex problem, in this work we focus on
two of the most basic hierarchical levels of the structure of a handwritten 
document and its corresponding DLA sub-problems, namely baseline detection and 
region segmentation. Nevertheless, we do 
not address the remaining sub-problems such as the reading order or other
relations between the elements of the document.

Limiting the DLA to these two sub-tasks allows us to address it 
as an object segmentation problem on images~\cite{haralick1985}, similar 
to~\cite{lquiros2018,lquiros2019,prusty2019}.

Many architectures have been proposed to address the object segmentation problem
on natural images using deep learning. We recommend~\cite{zhao2019} for a
comprehensive review of the most promising architectures. 

In view of
the good results proved by Mask-RCNN architecture~\cite{he2017} on natural images,
region segmentation on printed documents~\cite{zhong2019,saha2019} and historical 
documents~\cite{prusty2019,studer2019}, in this paper we analyze its effectiveness
on multi-task document layout analysis on complex handwritten historical documents.

Mask-RCNN is a multi-stage architecture specifically designed for multi-task 
problems. First, a \emph{backbone} network is used to extract a good 
feature representation of the input data. Then a \emph{Region Proposal Network}
is used to generate a set of rectangle boxes which likely contain
objects of interest. Finally, a \emph{Prediction Heads} network is used to perform
a set of tasks on the objects of interest, for instance classification, bounding 
box regression and mask generation.

Here we briefly describe Mask-RCNN architecture, while implementing details are 
presented in Sec.~\ref{sec:exp}

\paragraph{\emph{Backbone}:} The \emph{backbone} network is composed of 
a convolutional network that extracts features from the input image. Commonly,
the \emph{backbone} is composed of blocks of ResNet~\cite{he2016}, 
VGG~\cite{simonyan2014} or similar architectures that extract features at the 
scale of the input image. Also, it 
can be combined with a feature-pyramid network~\cite{lin2017} to extract features
at different hierarchical scales.

\paragraph{\emph{Region Proposal Network}:} It is another convolutional network
that scans the feature maps generated by the \emph{backbone} by an sliding window,
simultaneously
regressing region bounds and scoring the membership of each region to a set of objects
classes. On each location of the sliding window, $k$ region 
proposals are predicted, parameterized relative to a set of $k$ reference boxes
or anchors. Typically anchors span a range of scales and aspect ratios (e.g. 32,
64, 128 and 1:1, 1:2, 2:1).

Then each region proposal is filtered by its scores and a non-maximal suppression
algorithm
is used to handle overlaps. Finally, the region proposals that remain are 
called ``regions of interest'' (RoI) and passed to the next stage.

\paragraph{\emph{Prediction Heads}:} also called Prediction Branches, are a
set of neural networks (convolutional or not) designed to generate a task specific
prediction for each RoI generated on the previous stage. For instance, the most common
prediction heads for object segmentation are region labeling, region bounding box
regression and mask prediction. 

Region labeling classifies each RoI into a set of ``object classes'', for instance
in DLA, a set of classes could be text-line, paragraph and marginalia. On the
other hand, region bounding box regression maps each RoI to an adjusted box that 
better circumscribe the detected object. This new bounding box could 
differ from the anchor used to generate the RoI. Finally, the mask prediction head 
classifies each pixel of the RoI as belonging or not to the silhouette of the 
detected object. 

It is worth noting that the ability of the architecture to make predictions
for each RoI independently of the others, allows the model to naturally address
cases where the regions or objects touch or overlap each other. 
This is a major improvement over
Unet-like approaches~\cite{lquiros2018,ares2018,gruning2018} where overlapping 
objects are very difficult to handle.

\subsection{Baseline detection}
\label{sec:bl}
As mentioned before in this paper we aim to analyze the effectiveness of the Mask-RCNN
architecture to handle baseline detection and region segmentation on 
handwritten historical documents. While region segmentation is straightforwardly
addressed by Mask-RCNN architecture, baseline detection is not. 

The baselines of a document are normally represented as a piece-wise linear curve
and cannot be feed directly to Mask-RCNN framework. Instead we use use the 
polygon that enclosses each text line as the object to search for, in order to 
detect the baselines. If this polygon is not available on the ground-truth, it can
be easily generated as a fixed amount of pixels above and belong the baseline. 
Then for each ``text-line'' object detected by the model, we use the baseline 
detection algorithm presented in~\cite{lquiros2018} to obtain the baseline itself.

As counter intuitive this could sound\textemdash we convert
baselines to text lines, then to baselines and finally to text lines again in order
to feed any text recognition system \textemdash this methodology give us the 
versatility to handle ground-truth labeled using the cheaper baselines instead
of the cumbersome text line annotation~\cite{romero2015}.

\section{Experimental Setup}\label{sec:exp}

Our experiments are designed to analyze the performance of Mask-RCNN on complex 
handwritten documents. To that end, we experiment on different datasets, different
initialization methods and compare them to state-of-the-art methods for 
multi-task DLA.

\subsection{Datsets}
In order to estimate the effectiveness of the Mask-RCNN architecture on handwritten documents we 
select three publicly available datasets. Particularly, because we want to analyze the 
performance of the model in a multi-task set-up, all datasets are annotated not
only at baseline (text line) level, but a region level as well.

Two of the datasets are composed of only text related regions (paragraphs, 
marginalia, ...). This kind of dataset is expected to be very difficult for
any object detection model, because the boundary which separates a region and another
could be very subtle, specially on handwritten documents 
(See Fig.~\ref{fig:girona_page} and Fig.\ref{fig:vorau_layout}).
The other dataset (VORAU-253) is a music manuscript. It was selected in order
to analyze the performance of the model on
handwritten documents where the regions contains more than just text. 

\subsubsection{\textit{Oficio de Hipotecas de Girona}:}\label{sec:girona}
\begin{figure}[ht]
    \centering
    \begin{tikzpicture}
        \node[anchor=south west,inner sep=0] (image) at (0,0) {
            \subfloat[]{{\includegraphics[width=0.45\linewidth]{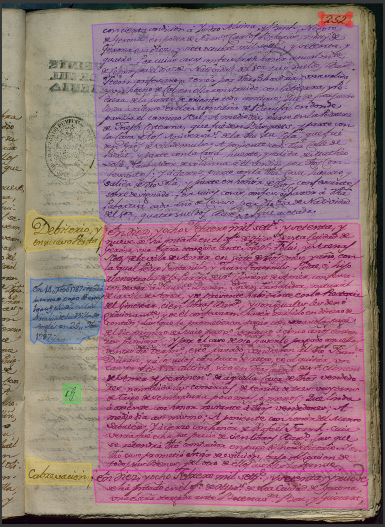}}}
            \subfloat[]{{\includegraphics[width=0.45\linewidth]{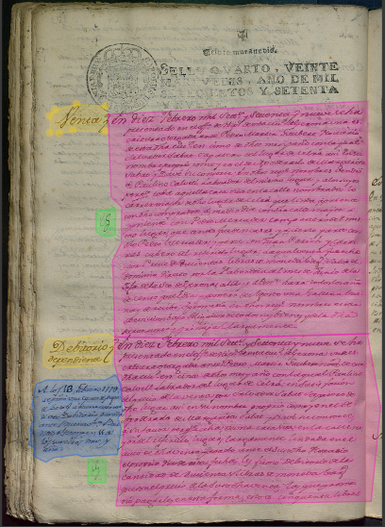}}}
        };
        \begin{scope}[x={(image.south east)},y={(image.north west)}]
            \draw[black,thick,rounded corners] (0.64,0.35) rectangle (1.0,0.45);
        \end{scope}
    \end{tikzpicture}
        \caption{Examples of a page of the OHG dataset. (a) All region classes that can be in an image.
        (b) An example of a subtle boundary between two \textit{par} regions (black bbox). Red=\textit{pag},
        yellow=\textit{tip}, violet=\textit{pac},
        pink=\textit{par}, blue=\textit{not},
        green=\textit{nop}.
        Better seeing in color.}
        \label{fig:girona_page}
\end{figure}

the manuscript (OHG) 
is composed of hundreds of 
thousands of  notarial deeds from the 18-19 centuries
(1768-1862)~\cite{Quiros2018}. Sales, redemption of censuses, inheritance and 
matrimonial chapters are among the most common documentary typologies in the 
collection. This collection exhibit a 
relatively complex layout, composed of six relevant region types; namely: 
\textit{pag}, \textit{tip}, \textit{par}, \textit{pac}, \textit{not}, \textit{nop}.
An example is depicted in Fig.~\ref{fig:girona_page}.

In this work we use the same training and test set defined in~\cite{lquiros2018},
composed of 300 images for training and 50 for test.
The main characteristics of this dataset are 
summarized on Table~\ref{tab:corpus}.

\begin{table}[ht]
    \setlength{\tabcolsep}{3pt}
    \caption{Main characteristics of the OHG dataset.}
    \begin{center}
        \begin{tabular}{lrrrrrr}
            \hline
            Region&\multicolumn{3}{c}{\#Regions} & \multicolumn{3}{c}{\#Lines} \\
            \cmidrule(lr){2-4} \cmidrule(lr){5-7}
            Name         & Train & Test & Total & Train & Test & Total\\
            \hline
            \textit{par} & 422 & 72 & 494 & 7716 & 1250  & 8966 \\ 
            \textit{pac} & 239 & 34 & 273 & 3352 &  554  & 3906 \\
            \textit{tip} & 418 & 72 & 490 &  449 &   75  & 524 \\
            \textit{pag} & 144 & 28 & 172 &  141 &   28  & 169 \\
            \textit{nop} & 184 & 37 & 221 &  180 &   37  & 217 \\
            \textit{not} & 18  & 3  & 21  &  112 &   16  & 128 \\
            \hline
        \end{tabular}
        \label{tab:corpus}
    \end{center}
    \setlength{\tabcolsep}{6pt}
\end{table}

\subsubsection{Bozen:}
this dataset consists of a subset of documents from the Ratsprotokolle collection
composed of minutes of the council meetings held from 1470 to 1805 (about 30\,000
pages)\cite{sanchez2016}. The dataset text is written in Early Modern German by an
unknown number of writers.
The public dataset is composed of 400 pages (350 for training and 50 for 
validation); most of the pages consist of two or three regions with many 
difficulties for line detection and extraction.

\begin{figure}[ht]
    \centering
        \subfloat[]{{\includegraphics[width=0.45\linewidth]{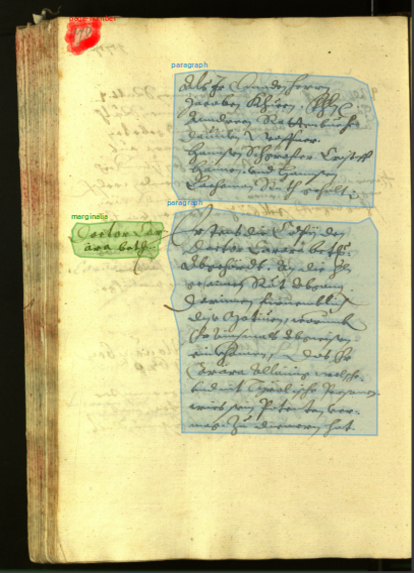}}}
        \subfloat[]{{\includegraphics[width=0.45\linewidth]{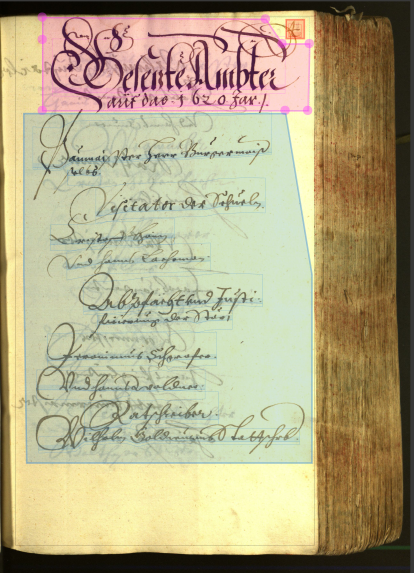}}}
        \caption{Examples of a page of the Bozen dataset. (a) Multi paragraph image.
        (b) A page with a header. Red=\textit{page-number}, green=\textit{marginalia}, 
        blue=\textit{paragraph}, pink=\textit{heading}. 
        Better seeing in color.}
        \label{fig:bozen_page}
\end{figure}

The ground-truth of this dataset is available in PAGE format (see~\cite{sanchez2016})
and annotated at baseline level and region level. 
Regions have been labeled using four different region types, namely: \textit{page-number}, 
\textit{marginalia}, \textit{paragraph} and \textit{heading}. 
The main characteristics are summarized on Table~\ref{tab:bozen}
and an example shown in Fig.~\ref{fig:bozen_page}.  

\begin{table}[ht]
    \setlength{\tabcolsep}{3pt}
    \caption{Main characteristics of the Bozen dataset.}
    \begin{center}
        \begin{tabular}{lrrrrrr}
            \hline
            Region&\multicolumn{3}{c}{\#Regions} & \multicolumn{3}{c}{\#Lines} \\
            \cmidrule(lr){2-4} \cmidrule(lr){5-7}
            Name         & Train & Test & Total & Train & Test & Total\\
            \hline
            \textit{page-number} & 350  & 50 & 400 & 350  & 50  & 400  \\  
            \textit{paragraph}   & 788  & 87 & 875 & 7118 & 894 & 8012 \\
            \textit{marginalia}  & 454  & 45 & 499 & 877  & 99  & 976  \\
            \textit{heading}     & 10   & 0  & 10  & 22   & 0   & 22   \\
            \hline
        \end{tabular}
        \label{tab:bozen}
    \end{center}
    \setlength{\tabcolsep}{6pt}
\end{table}

\subsubsection{VORAU-253:} 
\begin{figure}[ht]
    \centering
    \begin{tikzpicture}
        \node[anchor=south west,inner sep=0] (image) at (0,0) {
            \subfloat[]{{\includegraphics[width=0.45\linewidth]{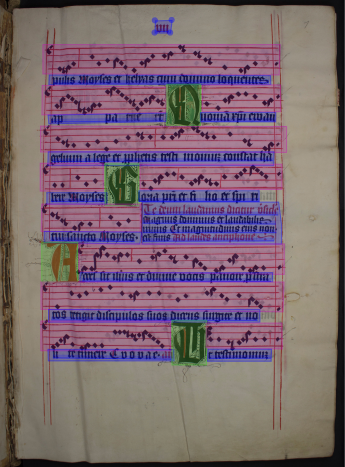}}}
            \subfloat[]{{\includegraphics[width=0.45\linewidth]{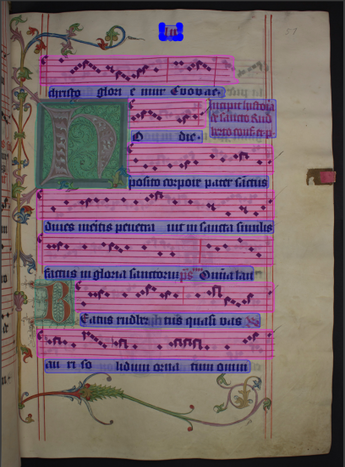}}}
        };
        \begin{scope}[x={(image.south east)},y={(image.north west)}]
            \draw[red,thick,rounded corners] (0.18,0.47) rectangle (0.25,0.60);
        \end{scope}
    \end{tikzpicture}
        \caption{Examples of a page of the VORAU-253 dataset. (a) A complex page
        with very subtle boundary between three different \textit{lyrics} regions (red bbox).
        (b) An example of a very decorated image. Green=\textit{drop-capital}, 
        pink=\textit{staff}, blue=\textit{lyrics}. 
        Better seeing in color.}
        \label{fig:vorau_layout}
\end{figure}

is a music manuscript  of the Vorau Abbey library, 
referred to as Cod. 253,
which was provided by the Austrian Academy of 
Sciences. It is written in German Gothic notation and dated around year 1450.
This manuscript is interesting because of the complexity of its layout, where staff, 
text and decorations are intertwined to compose the structure of the 
document (see Fig.~\ref{fig:vorau_layout}).

In this work we carried out our experiments using the same training
and test subsets as in~\cite{lquiros2019}, composed of
128 images for training and 100 for test.
The dataset contains three different layout elements: 
\begin{itemize}
    \item \textit{staff}: Represents the regions that contains a set of 
        horizontal lines and spaces, each one representing a different musical
        pitch.
        This region type does not contain text lines. Hence, no baselines.
    \item \textit{lyrics}: words that are sung, which appear
        below their corresponding staff, and other text in the document.
    \item \textit{drop-capital}: a decorated letter that might appear at the 
        beginning of a word or text line.
\end{itemize}

Main statistics of the dataset are presented in Table~\ref{tab:vorau}, and an 
example of an annotated page can be seeing in Fig.~\ref{fig:vorau_layout}.

\begin{table}[ht]
    \setlength{\tabcolsep}{3pt}
    \caption{Main characteristics of the VORAU-253 dataset.}
    \begin{center}
        \begin{tabular}{lrrrrrr}
            \hline
            Region&\multicolumn{3}{c}{\#Regions} & \multicolumn{3}{c}{\#Lines} \\
            \cmidrule(lr){2-4} \cmidrule(lr){5-7}
            Name         & Train & Test & Total & Train & Test & Total\\
            \hline
            \textit{drop-capital} & 336  & 232  & 568  & ---  & ---  & ---  \\
            \textit{staff}        & 1194 & 919  & 2113 & ---  & ---  & ---  \\
            \textit{lyrics}       & 1379 & 1042 & 2403 & 1628 & 1215 & 2843 \\
            \hline
        \end{tabular}
        \label{tab:vorau}
    \end{center}
    \setlength{\tabcolsep}{6pt}
\end{table}

\subsection{Evaluation Measures}
To the best of our knowledge, there is no common meaningful evaluation measure 
able to assess the results obtained in both tasks jointly, therefore we present a 
set of metrics for each task.

\paragraph{Region segmentation:} 
we report metrics from semantic segmentation and scene parsing evaluations as
presented in \cite{long2015}:

Mean Jaccard Index (mIoU): %
\begin{equation}
    \mathrm{mIoU} = 
\frac{1}{K}\sum_{i=1}^{K} \frac{\eta_{ii}}{(\tau_i+\sum_{j=1}^{K} \eta_{ji}-\eta_{ii})}
\end{equation}

Frequency weighted Jaccard Index (f.w.IoU): %
\begin{equation}
    \mathrm{f.w.IoU} = (\sum_{\kappa=1}^{K} \tau_\kappa)^{-1} \sum_{i=1}^{K} \frac{\tau_i \eta_{ii}}{\tau_i+\sum_{j=1}^K \eta_{ji}-\eta_{ii}}
\end{equation}
where $K$ is the number of different classes in the task, 
$\eta_{ij}$ is the number of pixels of class $i$ predicted to belong to 
class $j$, 
and $\tau_i$ the number of pixels of class $i$.
Even though we present both metrics, we consider Mean Jaccard Index the most
relevant one for the region segmentation task.

\paragraph{Baseline detection:}
we report precision (P), recall (R) and its harmonic mean (F1) measures as 
defined specifically for this kind of problem in \cite{Gruning2017}. Tolerance 
parameters are set to default values in all experiments. Notice that precision 
and recall are defined respect to the segments of the so called 
``normalized baseline'' and not directly to the whole baseline
(see \cite{Gruning2017} 
for details about measure definition, tolerance values and implementation 
details).

\subsection{System setup}
As explained before we use the Mask-RCNN architecture in all experiments. As 
\emph{backbone} we use the first four layers of ResNet50~\cite{he2016} 
combined with FPN~\cite{lin2017}. 
In order to evaluate how dependent Mask-RCNN is to the amount and type of training 
data available, the model has been trained using the following strategies:
\begin{itemize}
    \item[]Mask-RCNN-s: a model is trained from scratch for each dataset
        using only the data available of each respective  dataset.
    \item[]Mask-RCNN-A: a single model is trained from scratch using
        all the data available on the three datasets together.
    \item[]Mask-RCNN-I: on this case the weights of the \emph{backbone} were
        initialized using ImageNet dataset~\cite{deng2009}. Then, a model is
        fine-tuned for each dataset using the data data available of each 
        respective dataset.
    \item[]Mask-RCNN-I+A: like in the Mask-RCNN-I case the weights of the \emph{backbone}
        were initialized using ImageNet dataset. But, instead of three different models
        a single model is fine-tuned using all the data available on the three 
        datasets together. 
    \item[]Mask-RCNN-P: the weights of the \emph{backbone} were
        initialized using PubLayNet~\cite{zhong2019} dataset. Then, a model is
        fine-tuned for each dataset using the data data available of each 
        respective dataset.
    \item[]Mask-RCNN-I+P: like in the Mask-RCNN-I case the weights of the \emph{backbone}
        were initialized using ImageNet dataset. Then, the model is fine-tuned using 
        PubLayNet dataset. Finally a model is fine-tuned for each 
        dataset using the data data available of each respective dataset.
\end{itemize}

\subsubsection{Training:} we fine-tuning the models through $90K$ iterations using
the respective train set. Initial 
learning rate were set to 0.02 with a multi-step learning rate scheduler at $40K$,
$60K$ and $80K$ iterations.

In the case of models trained from scratch the initial learning rate were reduced
to $0.001$ in order to prevent the model from diverge at the beginning of the 
training process.

In all cases, RPN anchors are restricted to sizes in $\{32, 64, 128, 256, 512\}$ and
aspect-ratios in $\{1:1, 1:2, 2:1\}$, with a maximum number of region proposals set to 
$k=1000$ per image.

We use Stochastic Gradient Descent optimizer and a loss function composed of 
categorical 
cross entropy loss ($\mathcal{L}_{\mathrm{RPN}}$ and $\mathcal{L}_\mathrm{R}$) for RPN 
classification and region 
classification branch, smooth L1 loss $\mathcal{L}_{\mathrm{bb}}$ for bounding box 
prediction and binary cross entropy loss $\mathcal{L}_{\mathrm{mask}}$ for mask 
prediction. The total loss is a weighted combination of these losses, i.e.
$\mathcal{L} = \lambda_{\mathrm{RPN}}\mathcal{L}_{\mathrm{RPN}} + \lambda_{\mathrm{R}}\mathcal{L}_\mathrm{R} + 
\lambda_{\mathrm{bb}}\mathcal{L}_{\mathrm{bb}} + \lambda_{\mathrm{mask}}\mathcal{L}_{\mathrm{mask}}$. In our 
experiments we set $\lambda_{\mathrm{RPN}}=\lambda_{\mathrm{R}}=\lambda_{\mathrm{bb}}=\lambda_{\mathrm{mask}}=1$.

Batch size is set to 4 images in order to keep the model under hardware memory 
restrictions (we use only one Nvidia GeForce GTX 1080 GPU and 15GB of RAM). 

\subsubsection{Inference:} during inference we keep the number of RoI after 
non-maximal suppression to $1000$, then only the best $m$ RoI sorted by score
are feed to the \emph{Prediction Heads} stage, where $m$ is selected based
on the maximum number of objects in a training image $n$ for each dataset as 
$m=\max(100,n+50)$.

After network inference the results are post-processed to remove regions with a 
probability to belong to any class lower than $0.5$. Then, normal regions are 
separated from ``text-line'' to apply the baseline detection algorithm 
as discussed in Sec.~\ref{sec:bl}. Finally, ``text-lines'' and their respective baselines are
inserted hierarchically into the region with maximum IoU.

\section{Results}
\label{sec:results}

Results of basic training configurations (Mask-RCNN-\{s,I\}) 
along with state-of-the-art results available are presented for comparison in
Table~\ref{tab:sota}.

\begin{table*}[htb]
    \centering
    \caption{Multi-task DLA results for the basic
    training configurations proposed on the three datasets. State-of-the-art results
    also provided for comparison. Baseline detection metric (F1) and region 
    segmentation metric (mIoU) are presented in percentage form.
    In all cases the higher the better.}\label{tab:sota}
    \begin{tabular}{l|cccccc}
        \hline
        \multirow{2}{*}{Model} & \multicolumn{2}{c}{OHG} & \multicolumn{2}{c}{Bozen} & \multicolumn{2}{c}{VORAU-253}\\
                               & ~F1~   & ~mIoU~  & ~F1~   & ~mIoU~  & ~F1~   & ~mIoU~  \\
        \hline
        Quir\'os~\cite{lquiros2018}         & 96.6 & 80.0 & 96.6 & 84.5 & -- & -- \\
        Kiessling~\cite{kiessling2020}      & 91.4 & 48.6 & 94.2 & 81.0 & -- & -- \\
        Gr\"uning el al.~\cite{gruning2018} & -- & -- & 96.8 & -- & -- & -- \\
        Quir\'os et al.~\cite{lquiros2019}  & -- & -- & -- & -- & 96.0 & 87.0 \\
        Mask-RCNN-s                         & 93.6 & 81.3 & 84.8 & 71.5 & 30.4 & 33.2 \\
        Mask-RCNN-I                         & \textbf{96.7} & \textbf{85.5} & \textbf{ 97.9} & \textbf{86.8} & \textbf{98.4} & \textbf{89.5} \\
        \hline
    \end{tabular}
\end{table*}

In general Mask-RCNN is able to obtain very good results in all three
datasets, surpassing state-of-the-art methods for Mask-RNN-I, where the 
\emph{backbone} weights were initialized using a pre-trained model. 

Models trained from scratch using only the training data of the dataset itself
(Mask-RCNN-s) performs worse than the pre-trained counterparts. Likewise, 
those models are not able beat state-of-the-art models such 
as~\cite{lquiros2018,gruning2018,lquiros2019}, which are fairly comparable to
Mask-RCNN-s because they use only the training data of each dataset.

\begin{figure}[htb]
    \centering
    \begin{tikzpicture}
        \node[anchor=south west,inner sep=0] (image) at (0,0) {
            \subfloat[Ground-truth]{{\includegraphics[width=0.33\linewidth]{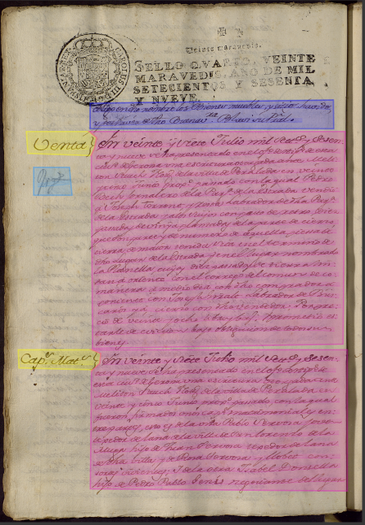}}}
            \subfloat[\cite{lquiros2018} model]{{\includegraphics[width=0.33\linewidth]{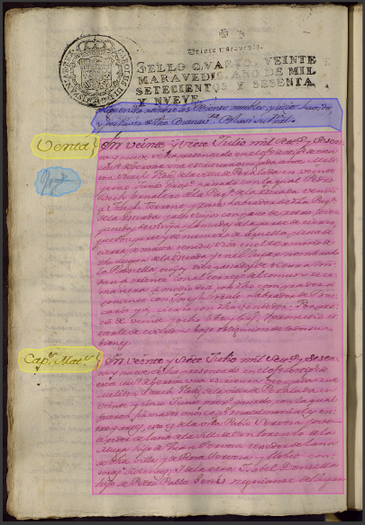}}}
            \subfloat[Mask-RCNN model]{{\includegraphics[width=0.33\linewidth]{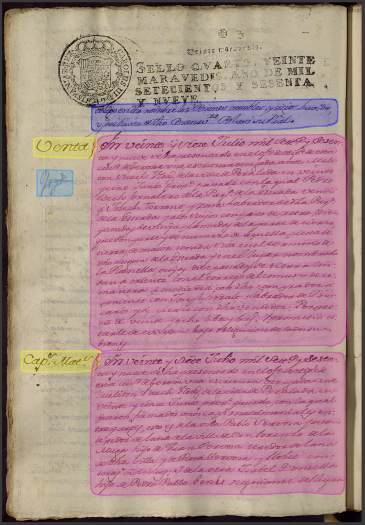}}}
        };
        \begin{scope}[x={(image.south east)},y={(image.north west)}]
            \draw[black,thick,rounded corners] (0.0,0.3) rectangle (1.0,0.42);
        \end{scope}
    \end{tikzpicture}
    \caption{Example of a page where two regions touch each other, pink regions
    inside the black bounding box area should be separated. 
    a) Shows the ground-truth where both regions touch. 
    b) \cite{lquiros2018} based model, regions are merged into a single one. 
    c) Mask-RCNN based model, touching regions are well handled.
    Better seeing in color.}
    \label{fig:overlap_example}
\end{figure}

On the other hand, pre-trained model surpasses state-of-the-art models in all cases.
It is important to notice that the region segmentation metric (mIoU) is 
specially improved by Mask-RCNN-I model. This is because Mask-RCNN is able to 
handle overlapping and touching regions, while previous models could not. For instance,
in the Fig.~\ref{fig:overlap_example} we can see an example of two touching regions 
that are
well handled by a Mask-RCNN model, but merged into a single one by a model
trained using the architecture proposed in~\cite{lquiros2018}.

Furthermore, in Table~\ref{tab:results} all trained models are presented for 
comparison\footnote{Mask-RCNN-\{s,I\} results are presented in Table~\ref{tab:sota}
and again in Table~\ref{tab:results} to facilitate comparison between models.}. 
Similarly to Mask-RCNN-I model, most of the pre-trained models achieve state-of-the-art results or better.

\begin{table*}[htb]
    \centering
        \caption{Multi-task DLA results for all the training strategies and datasets.
        Metrics are presented in percentage form, also in
        all cases the higher the better. Higher result of each dataset is 
        highlighted, but it does not mean that is statistically superior than others.}
        \label{tab:results}
    \begin{tabular}{l|lccc}
        \hline
        \multirow{2}{*}{Dataset} & \multirow{2}{*}{Model} & Baselines & \multicolumn{2}{c}{Regions} \\ 
                                 &                          & ~F1~   & ~mIoU~  & ~f.w.IoU~ \\
        \hline
        \multirow{6}{*}{OHG}    %
                                & Mask-RCNN-s                         & 93.6  & 81.3  & 90.7  \\
                                & Mask-RCNN-A                       & 75.8   & 58.0   & 70.0  \\
                                & Mask-RCNN-I                       & 96.7  & 85.5  & 92.7  \\
                                & Mask-RCNN-I+A                     & \textbf{96.8}  & \textbf{85.5} & \textbf{92.9}  \\
                                & Mask-RCNN-P                       & 94.1  & 81.1  & 91.4  \\
                                & Mask-RCNN-I+P                     & 96.7  & 84.8  & 92.8  \\
        \hline
        \multirow{6}{*}{Bozen}    %
                                  & Mask-RCNN-s                           & 84.8  & 71.5  & 83.7  \\
                                  & Mask-RCNN-A                         & 75.8  & 67.2 & 82.8  \\
                                  & Mask-RCNN-I                         &\textbf{ 97.9} & 86.8 & \textbf{92.5}  \\
                                  & Mask-RCNN-I+A                       & 97.5  & \textbf{87.0} & 92.3  \\
                                  & Mask-RCNN-P                         & 94.8  & 84.2  & 90.0  \\
                                  & Mask-RCNN-I+P                       & 97.1  & 85.7  & 92.3  \\
        \hline
        \multirow{6}{*}{VORAU-253}  %
                                    & Mask-RCNN-s                             & 30.4   & 33.2   & 54.9  \\
                                    & Mask-RCNN-A                           & 72.6   & 51.9   & 62.3  \\
                                    & Mask-RCNN-I                           & \textbf{98.4} & 89.5 & 91.4   \\       
                                    & Mask-RCNN-I+A                         & 98.1  & 90.4  & 92.4   \\       
                                    & Mask-RCNN-P                           & 97.1  & 88.1  & 90.3   \\       
                                    & Mask-RCNN-I+P                         & 98.1  & \textbf{90.7} & \textbf{92.6} \\       
        \hline
    \end{tabular}
\end{table*}

With respect to the models initialized using the PubLayNet dataset (Mask-RCNN-P), 
which is composed of printed documents, we observe no improvement of the results with respect 
to a general initialization using the ImageNet dataset (composed of natural images).
Even in the cases where a tandem initialization was used (Mask-RCNN-I+P) the 
improvement observed is very modest.

Mask-RCNN-A and Mask-RCNN-I+A are interesting cases. In 
both cases all three datasets were merged into a single bigger dataset. Recall
each dataset is composed of a disjoint set of regions. 

For Mask-RCNN-A we observe a degradation of the results with respect to 
the models where the datasets are used separately. This is mainly because some of
the RoI's generated by the model belong to a class of another dataset (e.g. a 
region of type \textit{staff} is assigned to a page of the OHG dataset). Besides, VORAU-253
dataset is an exception, mainly because the Mask-RCNN-s results are very low due
the lack of training data (only 128 images).

In contrast, Mask-RCNN-I+A is able to obtain very good results (albeit statistically
indifferent from Mask-RCNN-I) without any mislabeling observed between datasets.
This is an important result that tell us about the versatility of the model, since we can
train a single model and feed it with all incoming data without splitting it into 
the dataset it belongs to\footnote{This a is a important problem in libraries
and archives, where documents are often placed in boxes with unsorted mixed material.}. 
Also, from the point of view of processing large collections is important to 
reduce the number of models (i.e. the amount of memory used for them) in order to
process the data efficiently and without compromising the results obtained.

\section{Conclusions}
\label{sec:conc}

In this work we analyze the effectiveness of models based on Mask-RCNN to handle
DLA in a multi-task manner (baseline detection, region segmentation and region 
classification). In
general we have found that Mask-RCNN models need a lot of training data to achieve 
state-of-the-art results, however transfer learning helps to overcome this 
drawback. We obtain very good results using pre-trained models with the ImageNet
and PubLayNet datasets, nevertheless better results were obtained using the 
first one.

Furthermore, experimental evidence indicates that the Mask-RCNN model can handle 
several datasets at the same time without compromising the effectiveness of the 
model. This important experimental result shows the versatility of the model
and how it could be used to handle large heterogeneous collections.

In this work we analyze only one architecture based on region proposals. Future
work will consider additional architectures to analyze the relationship 
between complexity and effectiveness. Equally important is to modify the 
\emph{Prediction Heads} module to handle the relationship between layout elements, for 
instance the reading order and entity linking.

\section{Reproducibility}
\label{sec:rep}
Details required to reproduce our experiments are given below.

\paragraph{Source code:} we built upon
Detectron2\footnote{\url{https://github.com/facebookresearch/detectron2}} an 
open-source implementation of the Mask-RCNN architecture. Consequently, all 
modifications made, along with the configuration files to run the experiments, 
are publicly available in our online repository \repoURL.

\paragraph{Dataset splits:} we used the same dataset splits defined in previous 
works in order to keep comparability between experiments. OHG data splits can be found
in the repository given in~\cite{lquiros2018}\footnote{\url{https://github.com/lquirosd/P2PaLA/tree/master/egs/OHG}}.
For the Bozen dataset we used the same splits defined by the authors of the
dataset~\cite{sanchez2016}. Finally, we adopted the same data splits defined for 
the VORAU-$253$ dataset in~\cite{lquiros2019}, which is available in our 
repository as well\footnote{\repoURLdata}.

\paragraph{Hardware:} we ran all our experiments on one GPU (Nvidia GeForce GTX 
1080) and an Intel(R) Core(TM) i3-6100 CPU with 16GiB of memory. The hardware setup 
is important on this case because the Mask-RCNN architecture used in the experiments
is very memory demanding. For instance, datasets with very high number of \emph{regions
of interest} might not be possible to handle using this hardware setup.

\section*{Acknowledgements}
\acknowledge
 \bibliographystyle{splncs04}
 \bibliography{main}
\end{document}